\documentclass{article}

\usepackage[numbers]{natbib}
\usepackage{PRIMEarxiv}
\usepackage{listings}
\usepackage[utf8]{inputenc} 
\usepackage[T1]{fontenc}    
\usepackage{hyperref}       
\usepackage{url}            
\usepackage{booktabs}       
\usepackage{amsmath,amssymb,amsfonts}
\usepackage{algorithm}
\usepackage{comment}
\usepackage{amsmath}
\usepackage{algpseudocode}
\usepackage{algorithm}
\usepackage{nicefrac}       
\usepackage{microtype}      
\usepackage{fancyhdr}       
\usepackage{graphicx}       
\graphicspath{{media/}}     
\usepackage{booktabs}
\usepackage{xcolor}
\title{NASP-T: A Fuzzy Neuro-Symbolic Transformer for Logic-Constrained Aviation Safety Report Classification
}

\author{
  Fadi Al Machot\thanks{Corresponding author: Fadi Al Machot (e-mail: fadi.al.machot@nmbu.no)} \\
  Faculty of Science and Technology (REALTEK) \\
  Norwegian University of Life Sciences (NMBU) \\
  1430 Ås, Norway
  \and
  Fidaa Al Machot \\
  Dresden International University (DIU) \\
  Freiberger Str.~37, 01067 Dresden, Germany \\
  fidaa.machot@gmail.com
}

\begin{document}
\maketitle

\begin{abstract}
Deep transformer models excel at multi-label text classification but often violate domain logic that experts consider essential—an issue of particular concern in safety-critical applications. We propose a hybrid neuro-symbolic framework that integrates Answer Set Programming (ASP) with transformer-based learning on the Aviation Safety Reporting System (ASRS) corpus. Domain knowledge is formalized as weighted ASP rules and validated using the Clingo solver.\footnote{Clingo: \url{https://potassco.org/clingo/}} These rules are incorporated in two complementary ways: (i) as rule-based data augmentation, generating logically consistent synthetic samples that improve label diversity and coverage; and (ii) as a fuzzy-logic regularizer, enforcing rule satisfaction in a differentiable form during fine-tuning. This design preserves the interpretability of symbolic reasoning while leveraging the scalability of deep neural architectures. We further tune per-class thresholds and report both standard classification metrics and logic-consistency rates. Compared to a strong Binary Cross-Entropy (BCE) baseline, our approach improves micro- and macro-F1 scores and achieves up to an 86\% reduction in rule violations on the ASRS test set. To the best of our knowledge, this constitutes the first large-scale neuro-symbolic application to ASRS reports that unifies ASP-based reasoning, rule-driven augmentation, and differentiable transformer training for trustworthy, safety-critical NLP.

\end{abstract}

\section{Introduction}

Aviation safety remains a critical priority as global air traffic continues to grow. Despite major technological advances, human factors, environmental conditions, and system complexity still contribute to incidents that pose risks to flight safety. To mitigate these risks, reporting systems such as the \textit{Aviation Safety Reporting System (ASRS)} were established in 1976 by NASA as a voluntary, confidential platform where pilots, controllers, and other aviation personnel can submit safety incident reports. Today, ASRS contains over a million narratives describing anomalies ranging from equipment failures to runway incursions, making it one of the richest corpora for studying aviation safety events \cite{andrade2023safeaerobert, ahmed2010multi}.

Processing these reports is challenging due to their free-text nature, domain-specific jargon, and the fact that a single report often corresponds to multiple overlapping categories (\textit{multi-label classification}). For example, a single ASRS entry may simultaneously describe a ``Runway Incursion,'' ``Non-Adherence to Legal Separation,'' and ``Communication Breakdown'' \cite{ahmed2010multi}. Conventional single-label or binary classifiers fail to capture this complexity, while even advanced deep learning approaches struggle with the noisy, sparse, and highly technical language of aviation narratives.

Recent efforts have sought to improve automatic classification and information extraction from ASRS. SafeAeroBERT \cite{andrade2023safeaerobert}, a domain-specific language model pre-trained on ASRS and National Transportation Safety Board (NTSB) reports, demonstrated improved performance over general-purpose BERT models in classifying safety events. 

However, a key limitation of existing approaches is that they do not incorporate \textit{safety rules or logical dependencies} into the learning process. Aviation safety knowledge is often expressed as conditional or causal statements (e.g., ``if an engine failure occurs, then a diversion is required''), yet current models treat classification as a purely statistical task. This omission leads to predictions that can violate known operational relationships, undermining interpretability and trust in high-stakes decision contexts. Addressing this gap requires combining the representational power of deep language models with the interpretability and structure of symbolic reasoning—a central goal of neuro-symbolic learning.

In this work, we present a hybrid \textbf{Neuro-Symbolic ASP-Constrained Transformer (NASP-T)} framework that integrates pre-trained transformer encoders with domain knowledge encoded as soft Answer Set Programming (ASP) rules. Using the \texttt{Clingo} solver, we formalize and validate these rules, which are then incorporated into the learning process through both symbolic preprocessing and differentiable training objectives. 

Our framework embeds safety knowledge at three complementary levels:
\begin{enumerate}
    \item \textbf{Rule-based Data Augmentation:} ASP-derived implications are used to generate logically consistent synthetic samples, enriching the training distribution and improving label balance.
    \item \textbf{Fuzzy ASP Regularization:} A differentiable fuzzy-logic loss softly penalizes violations of ASP-inspired rules during gradient-based optimization, encouraging logically coherent predictions.
    \item \textbf{Solver-based Validation:} During evaluation, the \texttt{Clingo} solver audits model predictions for rule compliance, producing interpretable logic-consistency metrics alongside standard classification measures.
\end{enumerate}

To the best of our knowledge, this is the first work to apply a neuro-symbolic framework that couples Clingo-based rule validation with differentiable fuzzy logic regularization for the ASRS aviation safety reports dataset.

\section{Related Work}
\subsection{Domain-specific Language Models}
General-purpose language models such as BERT have shown strong results on diverse NLP tasks, but their direct application to aviation text is limited by domain-specific jargon, abbreviations, and technical terminology. To address this, domain-adapted models have been proposed.

SafeAeroBERT \cite{andrade2023safeaerobert} was recently introduced as a safety-informed aerospace-specific variant of BERT, pretrained on over 400,000 ASRS and NTSB reports. The model is designed for downstream tasks such as document classification, named-entity recognition, relation extraction, and information retrieval. Initial evaluations show that SafeAeroBERT outperforms general-purpose BERT and SciBERT for certain classification tasks, particularly reports related to weather and procedural issues.

Similarly, Aviation-BERT \cite{jing2023bert} was developed as a domain-specific BERT variant pretrained on more than 500,000 ASRS and NTSB narratives. It has been fine-tuned for multi-label anomaly classification in ASRS reports, achieving higher F1 scores and lower Hamming loss compared to baseline BERT models. This demonstrates that aviation-specific pretraining enables better handling of technical terminology and improves classification accuracy in safety reporting systems.

\subsection{Neuro-Symbolic Learning and Constraint Integration}
A broader trend in AI research is the integration of neural networks with symbolic reasoning. Researchers in \cite{yu2023survey}, outline how combining statistical perception with logical reasoning can bridge the gap between pattern recognition and knowledge-driven inference. In this line, Semantic Loss \cite{ahmed2024semantic, arrotta2024semantic} encodes logical formulas into differentiable losses, encouraging neural models to respect known constraints. Other works adopt optimization-based formulations, such as Lagrangian Dual Methods \cite{rong2022lagrangian} and Augmented Lagrangian Relaxations \cite{kotary2024learning}, to enforce structured constraints during training. Physics-informed models, e.g., Computation Meets Physics \cite{basir2022physics}, show similar ideas for scientific modeling.

ASP-based reasoning has also gained traction. Answer Set Networks \cite{skryagin2024answer} and extensions of ASP to differentiable frameworks demonstrate how symbolic solvers like Clingo can interact with neural networks. Constraint satisfaction for NLP tasks has been studied in works like Constraint Reasoning in NLP \cite{jiang2022constraint}, which highlights the potential of hybrid methods for ensuring consistency in multi-label prediction.

While most existing neuro-symbolic approaches have been evaluated on synthetic or structured datasets such as physics simulations, few have tackled noisy, real-world text corpora like ASRS. Moreover, prior aviation NLP systems such as SafeAeroBERT~\cite{andrade2023safeaerobert} and AviationBERT~\cite{jing2023bert} rely solely on domain-specific pretraining without explicit reasoning constraints. In contrast, our method integrates \emph{symbolic safety rules} into transformer training through a differentiable fuzzy-logic regularizer derived from ASP. Unlike semantic loss formulations~\cite{ahmed2024semantic, arrotta2024semantic}, Lagrangian relaxations~\cite{kotary2024learning}, or differentiable ASP approximations~\cite{skryagin2024answer}, we explicitly ground our rules using the Clingo solver for validation and consistency checking. This enables the incorporation of richer safety constraints—such as causal implications and co-occurrence patterns—beyond mutual exclusions.

\section{Dataset and Rule Generation}
We evaluate our proposed framework using the \textit{Aviation Safety Reports Text Classification} dataset, originally released as part of the \textbf{SIAM 2007 Text Mining Competition} \cite{siam2007dataset}. The corpus is derived from NASA’s \textit{Aviation Safety Reporting System} (ASRS), a large collection of voluntary, de-identified incident reports submitted by aviation personnel. Each report narratively describes one or more operational anomalies or system failures that occurred during a flight. The competition dataset was designed for multi-label document classification, where the task is to assign each report to one or more problem categories (e.g., \textit{Engine Problem}, \textit{Cabin Pressure Problem}, \textit{Navigation Problem}). Each line in the dataset corresponds to a single ASRS report, identified by a unique document number followed by its text content. The data are provided in raw text format with accompanying label matrices defining the relevant categories.

 The data are publicly accessible and intended for research use. Since the original reports were produced under U.S. government programs, the dataset is considered public domain (U.S. Government Work). We adopt the common train/validation/test split and represent each instance by its narrative paired with a multi-hot label vector. 

Because these labels exhibit strong but noisy co-occurrence patterns, we derive \emph{soft implication rules} in two complementary ways. First, we compute conditional supports from the training set to propose candidate implications such as ``if \texttt{Engine Failure} then \texttt{Emergency Landing},'' and assign weights proportional to empirical confidence. Second, we incorporate domain knowledge about operational dependencies (e.g., communication issues typically co-occurring with ATC interaction) to supplement the statistical rules. Each rule is recorded in a compact 
\texttt{soft\_rule(premise, conclusion, weight)} format, then enriched by mapping category identifiers to human-readable labels. The enriched rules are automatically converted into Answer Set Programming (ASP) weak constraints.

A soft implication rule of the form
\[
A_i \;\Rightarrow\; B_j \;(w_{ij})
\]
expresses that whenever condition \(A_i\) holds, conclusion \(B_j\) should also hold, with an associated confidence weight \(w_{ij} \in (0,1]\).  
In the ASP formalism, such a rule is encoded as a \emph{weak constraint}:
\begin{lstlisting}[language=Prolog]
:~ holds("A_i"), not holds("B_j"). [w_ij@1,"A_i","B_j"].
violation("A_i","B_j") :- holds("A_i"), not holds("B_j").
\end{lstlisting}
This formulation assigns a penalty of weight \(w_{ij}\) whenever the premise is satisfied but the conclusion is not, enabling the solver to search for stable models that minimize total rule violations.
In this notation, the symbol \texttt{@} specifies the \emph{optimization priority level} of the weak constraint, indicating that penalties with lower levels (e.g., \texttt{@1}) are minimized before those with higher levels in the solver’s lexicographic optimization process.

\section{Methodology}
\begin{figure}[t]
    \centering
    \includegraphics[width=0.5\linewidth]{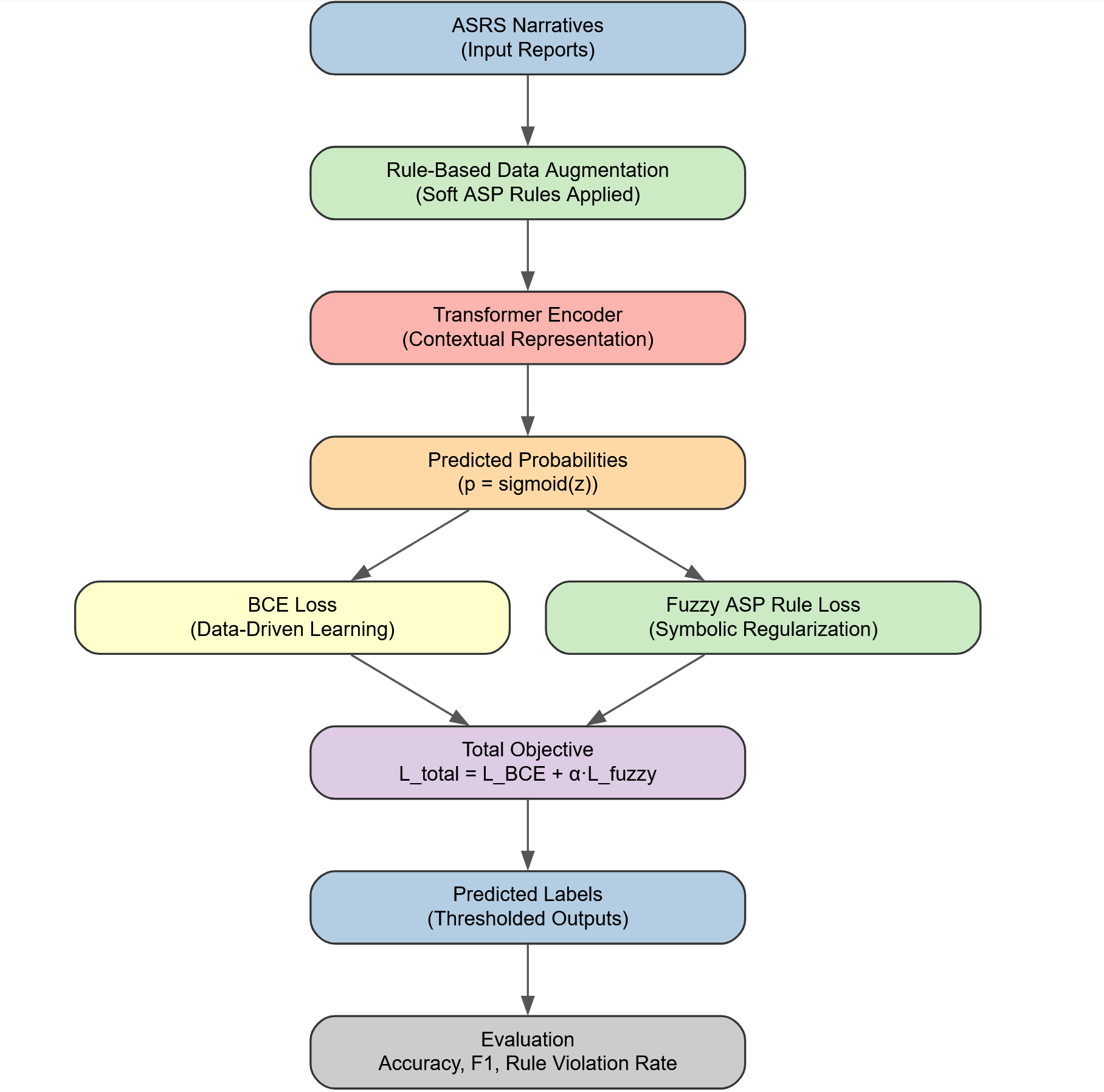}
\caption{
Overview of the proposed Neuro-Symbolic ASP-Constrained Transformer (NASP-T) framework. 
The pipeline integrates transformer-based text encoding with domain knowledge expressed as soft Answer Set Programming (ASP) rules. 
These rules are first used for logic-driven data augmentation, enriching the training set with rule-consistent examples, and later as a fuzzy ASP regularizer that penalizes logical inconsistencies during optimization. 
The BCE loss ensures data-driven learning, while the fuzzy ASP term enforces symbolic coherence. 
Together, they yield predictions that are both statistically accurate and logically consistent—crucial for safety-critical NLP tasks.
}
    \label{fig:neuro}
\end{figure}

Figure~\ref{fig:neuro} presents an overview of the proposed Neuro-Symbolic ASP-Constrained Transformer (NASP-T) framework. The approach integrates transformer-based text encoding with symbolic reasoning derived from domain-specific safety rules. In the preprocessing stage, these rules are first used for logic-driven data augmentation, enriching the training set with additional, rule-consistent examples. During training, the transformer predicts label probabilities across multiple safety categories and is optimized using two complementary objectives: the standard Binary Cross-Entropy (BCE) loss for statistical accuracy and a Fuzzy ASP loss that softly penalizes violations of logical dependencies. This dual learning process allows the model to internalize both empirical and symbolic structure. During inference, predictions are thresholded and evaluated not only with conventional multi-label metrics but also with a rule violation rate, ensuring outputs remain both accurate and logically coherent—crucial for safety-critical text analysis.

\subsection{Problem Setup}
We address a \emph{multi-label text classification} task over the Aviation Safety Reporting System (ASRS) corpus, where each report describes an aviation incident and can belong to multiple categories (e.g., \textit{Engine Failure}, \textit{Weather Issues}).  
Formally, let the input space be $\mathcal{X}$ and the label space $\mathcal{Y} = \{0,1\}^m$, where $m$ is the number of possible categories.  
Each training instance is a pair $(x_i, \mathbf{y}_i)$ where:
\[
x_i \in \mathcal{X} \text{ is the incident text, and } \mathbf{y}_i = [y_{i1}, y_{i2}, \dots, y_{im}] \in \{0,1\}^m
\]
is a multi-hot vector, where $y_{ij}=1$ if label $j$ applies to report $i$ and $0$ otherwise.

A transformer-based encoder $f_\theta$ (e.g., DeBERTa or DistilBERT) maps the input text $x_i$ into a hidden representation and outputs raw logits:
\[
\mathbf{z}_i = f_\theta(x_i) \in \mathbb{R}^m.
\]
After applying a sigmoid activation function, we obtain independent probabilities for each label:
\[
\mathbf{p}_i = \sigma(\mathbf{z}_i), \quad \text{where } p_{ij} = \frac{1}{1 + e^{-z_{ij}}}.
\]
Here, $p_{ij}$ represents the model's confidence that label $j$ should be active for input $x_i$.

\subsection{Baseline Loss: Binary Cross-Entropy (BCE)}
To handle class imbalance and independent label predictions, we employ the weighted Binary Cross-Entropy (BCE) loss:
\begin{equation}
\mathcal{L}_{\text{BCE}}(\theta) = 
-\frac{1}{N}\sum_{i=1}^{N}\sum_{j=1}^{m}
\Big[
w_j y_{ij} \log(p_{ij}) + (1 - y_{ij})\log(1 - p_{ij})
\Big].
\end{equation}
Here:
\begin{itemize}
    \item $N$ is the total number of training samples.
    \item $m$ is the number of labels.
    \item $w_j$ is a positive weight to counter class imbalance, defined as 
    $w_j = \frac{n_{\text{neg},j}}{n_{\text{pos},j} + 1e{-5}}$, 
    where $n_{\text{pos},j}$ and $n_{\text{neg},j}$ denote counts of positive and negative samples for label $j$.

\end{itemize}

Here, \(n_{\text{pos},j}\) and \(n_{\text{neg},j}\) denote the number of positive and negative training samples for label \(j\), respectively, where \(n_{\text{pos},j}\) counts instances in which the label is present and \(n_{\text{neg},j}\) those in which it is absent.
 
\subsection{Data Augmentation Using Logical Rules}
To improve consistency, we augment the training set using logical rules extracted from co-occurrence statistics and expert heuristics.  
Each rule is represented as a soft implication $(a \Rightarrow b, \omega)$ with weight $\omega \in (0,1]$, where $a$ and $b$ are label names (e.g., \texttt{Engine Failure} $\Rightarrow$ \texttt{Emergency Landing}).  

For each instance $(x_i, \mathbf{y}_i)$ in the training set, if $y_{ia}=1$ but $y_{ib}=0$, a new augmented instance $(x_i, \tilde{\mathbf{y}}_i)$ is added such that $\tilde{y}_{ib}=1$.  
This increases data coverage for logically implied labels and encourages the model to internalize these dependencies.

Formally, let $\mathcal{R}$ denote the rule set. The augmented dataset is:
\[
D^+ = D \cup \Big\{(x_i, \tilde{\mathbf{y}}_i) : (a\Rightarrow b, \omega)\in\mathcal{R}, y_{ia}=1, y_{ib}=0 \Big\}.
\]
\subsection{Fuzzy Rule Regularization}
While rule-based data augmentation improves label coverage, it cannot guarantee logical consistency during optimization.  
To explicitly enforce consistency, we define a differentiable \emph{fuzzy rule loss} grounded in fuzzy logic implications~\cite{hajek2001metamathematics}.

For each rule $(a \Rightarrow b, \omega)$ with corresponding predicted probabilities $p_a$ and $p_b$, the fuzzy satisfaction is:
\begin{equation}
\text{sat}(a \Rightarrow b) = \min(1, 1 - p_a + p_b),
\end{equation}
where a value of 1 denotes full satisfaction and lower values indicate partial or full violation.  
The corresponding violation score is:
\begin{equation}
v(a \Rightarrow b) = 1 - \text{sat}(a \Rightarrow b) = \max(0, p_a - p_b),
\end{equation}
which penalizes cases where the model predicts a premise with higher confidence than its expected consequence.

The overall fuzzy rule loss is the weighted mean violation across all rules and samples:
\begin{equation}
\mathcal{L}_{\text{fuzzy}}(\theta) =
\beta \cdot \frac{1}{|\mathcal{R}|}
\sum_{(a \Rightarrow b,\omega)\in\mathcal{R}}
\omega \, \mathbb{E}_i[\max(0, p_{i,a} - p_{i,b})],
\end{equation}
where:
\begin{itemize}
    \item $\beta$ (set via \texttt{fuzzy\_weight} in code) controls the strength of rule regularization relative to the BCE loss.
    \item $\omega$ represents the empirical confidence of each rule (e.g., 0.85 for “Engine Failure $\Rightarrow$ Emergency Landing”).
    \item $p_{i,a}$ and $p_{i,b}$ denote the predicted probabilities of the premise and conclusion for sample $i$.
\end{itemize}

\subsection{Relation to Symbolic Clingo Reasoning}
In the symbolic phase, we employ the \texttt{Clingo} Answer Set Programming (ASP) solver~\cite{gebser2014clingo} to formally encode and validate domain rules before integrating them into the learning process.  
Each implication rule of the form $a \Rightarrow b$ is represented in ASP as a weak constraint, allowing the solver to minimize rule violations and ensure logical consistency across candidate solutions.  
Although this symbolic reasoning provides a rigorous logical foundation, it cannot be directly embedded into neural optimization since ASP inference is inherently discrete and non-differentiable.  
To address this limitation, our framework converts these symbolic constraints into differentiable approximations through a fuzzy relaxation (described in Subsection~4.2), enabling the pretrained transformer to be fine-tuned end-to-end while remaining consistent with the logic validated by \texttt{Clingo}.  
This approach retains the interpretability and consistency of symbolic reasoning while leveraging the scalability and representational power of deep neural networks.

\subsection{Total Objective Function}
The final training objective combines the statistical BCE term with the logic-aware fuzzy loss:
\begin{equation}
\mathcal{L}_{\text{total}}(\theta) = 
\mathcal{L}_{\text{BCE}}(\theta) + 
\mathcal{L}_{\text{fuzzy}}(\theta).
\end{equation}
This encourages the model to balance predictive accuracy and rule consistency.  
In early epochs, $\mathcal{L}_{\text{BCE}}$ dominates, allowing the model to learn basic discriminative patterns; later, $\mathcal{L}_{\text{fuzzy}}$ regularizes the decision boundaries toward rule coherence.
\subsection{Rule Violation Metrics}
To complement traditional performance metrics (e.g., F1, Hamming loss), we explicitly measure the model’s logical consistency with respect to the domain rules.  
Let $\mathcal{R}$ denote the set of all soft implication rules of the form $(a \Rightarrow b)$, where $a$ is a \emph{premise} label and $b$ is a \emph{conclusion} label.  
Each test sample $i \in \{1, \ldots, N\}$ has predicted binary outputs 
$\hat{\mathbf{y}}_i = [\hat{y}_{i,1}, \hat{y}_{i,2}, \ldots, \hat{y}_{i,m}]$, 
where $\hat{y}_{i,j} \in \{0,1\}$ indicates whether category $j$ is predicted active for that sample after thresholding.

\paragraph{Total Rule Violations.}
We define the total number of rule violations across the test set as:
\begin{equation}
V = \sum_{i=1}^{N} \sum_{(a \Rightarrow b) \in \mathcal{R}} 
\mathbf{1}[\hat{y}_{i,a} = 1 \wedge \hat{y}_{i,b} = 0],
\end{equation}
where $\mathbf{1}[\cdot]$ is the indicator function that equals $1$ if its condition is true and $0$ otherwise.  
In words, a violation occurs when the model predicts the premise label $a$ (e.g., \textit{Engine Failure}) but fails to predict its expected conclusion $b$ (e.g., \textit{Emergency Landing}).  
This definition captures every contradiction between model predictions and the domain’s logical implications.

\paragraph{Violation Rate.}
Since some premises may occur more frequently than others, we also compute a normalized \emph{violation rate}:
\begin{equation}
\text{Rate} = 
\frac{V}{
\sum_{i=1}^{N} 
\sum_{(a \Rightarrow b) \in \mathcal{R}}
\mathbf{1}[\hat{y}_{i,a} = 1]
} \times 100,
\end{equation}
which measures the percentage of violated implications among all \emph{active premises}.  
For instance, if 10\% of all predicted \textit{Engine Failure} cases lack the corresponding \textit{Emergency Landing} prediction, then the violation rate for that rule is 10\%.

\paragraph{Document-Level Normalization.}
In reporting, we also present:
\begin{itemize}
    \item \textbf{Violations per Document:} $V / N$, the average number of rule violations per report.
    \item \textbf{Violations per 1{,}000 Documents:} $1000 \times (V / N)$, a scaled version for interpretability.
\end{itemize}

Together, these metrics provide a structural evaluation of model behavior.  
A lower violation count or rate indicates predictions that are more compliant with aviation safety logic.  In addition, we report both Micro-F1 and Macro-F1 to comprehensively evaluate classification performance.
Micro-F1 aggregates predictions across all labels, reflecting overall accuracy on frequent categories, while Macro-F1 averages per-label F1 scores, emphasizing balanced performance across both common and rare safety event types.
While F1 and accuracy quantify statistical fit, violation metrics reveal whether the model’s outputs are plausible under expert domain knowledge—essential for high-stakes applications such as incident classification.

We used the HuggingFace \texttt{transformers} library with the \texttt{AutoModelForSequenceClassification} architecture for $f_\theta$.  
Training hyperparameters included a learning rate of $3\times10^{-5}$, batch size of 8, and 6 epochs with early stopping based on validation F1.  
Each experiment was repeated with fixed random seeds for reproducibility.  
Data augmentation increased training size by 10–15\%, depending on rule coverage, and the fuzzy loss weight $\beta$ was tuned in $\{0.1, 0.5, 0.9\}$.

\section{Experimental Results}

\begin{table*}[ht]
\centering
\caption{
Test set performance comparing baseline BCE to BCE + fuzzy ASP regularization 
with rule-based data augmentation. Consistency is reported as total rule violations, 
violations per document, and violations per 1{,}000 documents. 
All results are computed on the 7{,}077-document ASRS test set.
}
\label{tab:main-results}
\begin{tabular}{lccccccc}
\toprule
Model  & Micro-F1 & Macro-F1 & Hamming & Violations & Viol./doc & Viol./1k \\
\midrule
\textbf{DistilBERT (BCE)} & 0.632 & 0.589 & 0.081 & 323 & 0.0456 & 45.6 \\
\textbf{DistilBERT + Fuzzy ASP (0.1)} & 0.629 & 0.587 & 0.081 & 332 & 0.0469 & 46.9 \\
\textbf{DistilBERT + Fuzzy ASP (0.5)} & 0.630 & 0.589 & 0.081 & 101 & 0.0143 & 14.3 \\
\textbf{DistilBERT + Fuzzy ASP (0.9)} & 0.632 & 0.588 & 0.080 & \textbf{44} & \textbf{0.0062} & \textbf{6.2} \\
\midrule
\textbf{DeBERTa-v3 (BCE)}  & 0.506 & 0.448 & 0.138 & 1159 & 0.1638 & 163.8 \\
\textbf{DeBERTa-v3 + Fuzzy ASP (0.1)}  & 0.548 & 0.487 & 0.119 & 797 & 0.1126 & 112.6 \\
\textbf{DeBERTa-v3 + Fuzzy ASP (0.5)}  & 0.325 & 0.067 & 0.183 & 0 & 0.0000 & 0.0 \\
\textbf{DeBERTa-v3 + Fuzzy ASP (0.9)}  & 0.296 & 0.073 & 0.223 & 0 & 0.0000 & 0.0 \\
\bottomrule
\end{tabular}
\end{table*}

Table~\ref{tab:main-results} summarizes the quantitative comparison between 
the baseline binary cross-entropy (BCE) and the proposed neuro-symbolic approach 
that augments training with fuzzy ASP regularization and rule-based data augmentation. 
The baseline DistilBERT achieves the highest micro-F1 score (0.632) but 
exhibits 323 logical violations, corresponding to approximately 0.046 violations per document. 
When ASP-based regularization is applied, rule violations drop significantly 
without a loss in F1 performance. With a fuzzy weight of 0.5, the violation count 
is reduced by nearly 70\%, and with stronger regularization (0.9), only 44 violations 
remain—representing a sevenfold improvement in logical consistency while 
maintaining comparable predictive accuracy. This shows that symbolic regularization 
can effectively constrain model behavior without compromising generalization.

In contrast, the larger DeBERTa-v3 model achieves a high accuracy 
in BCE-only training but produces substantially more rule violations (1,159 total). 
When fuzzy ASP constraints are added, violations decrease in proportion to the 
regularization strength. However, for overly strong weights ($\geq$ 0.5), 
the model underfits and yields lower F1 scores, confirming that an optimal 
balance between statistical learning and symbolic enforcement is necessary. 

These results highlight a core insight: neural models trained solely with 
statistical objectives (e.g., BCE) can achieve high predictive scores yet 
remain logically inconsistent. By integrating fuzzy ASP constraints, 
the model learns to align its latent decision boundaries with domain rules. 
In practice, this reduces implausible combinations of labels 
(e.g., predicting \texttt{Engine Failure} without \texttt{Emergency Landing}) 
and improves calibration on safety-critical conditions. 
The small F1 trade-off is outweighed by large gains in interpretability, 
trustworthiness, and auditability—key for safety-sensitive NLP applications.

Furthermore, the observed differences between DistilBERT and DeBERTa suggest 
that smaller architectures may benefit more from symbolic regularization, 
as their representational capacity leaves more room for rule-guided correction. 
This opens a new direction for lightweight, rule-aware models suitable for 
deployment in real-time aviation monitoring and safety management systems.

\section{Conclusion}
We introduced a hybrid neuro-symbolic framework for multi-label classification 
of aviation safety narratives, integrating fuzzy ASP rules into both 
data augmentation and model training. Our approach demonstrates that combining 
Answer Set Programming (ASP) with deep transformers yields models that are 
not only statistically strong but also logically consistent. 
By enforcing soft rules during learning and auditing rule compliance during 
evaluation, we significantly reduce violations of domain knowledge while 
preserving F1 performance.

Future work will explore integrating the \texttt{Clingo} solver directly 
into mini-batch training for dynamic constraint evaluation, 
adapting rule weights based on uncertainty calibration, and extending 
our logic layer to handle temporal and causal relations beyond document-level 
co-occurrence. The proposed NASP-T pipeline establishes a new paradigm for 
trustworthy, interpretable, and safety-aware NLP systems, 
paving the way for neuro-symbolic learning in other high-stakes domains 
such as healthcare, transportation, and risk intelligence.

\bibliographystyle{plain}
\bibliography{neuro}
\end{document}